\documentclass{fairmeta}

\usepackage{amsmath,amssymb,amsfonts,amsthm}
\usepackage{adjustbox}
\usepackage{array}
\usepackage{enumitem}
\usepackage{url}
\usepackage{tabularx}

\newcommand{\ctr}{\text{CTR}}

\title{Reasoning in Fewer Tokens: Concept-Normalized Multilingual Scratchpads and Reversible Model-Adaptive Reasoning Codes}

\author[1]{Rabimba Karanjai}
\author[1]{Suryabhan Singh Hada}
\author[1]{Hemanth Hegadehalli Madhavarao}
\author[1]{Qun Gu}
\author[1]{Libin N. George}
\author[1]{Wenhuan Sun}
\author[1]{Neha Monga}
\author[1]{Richard Williamson}
\author[1]{Prakhar Mehrotra}

\affiliation[1]{PayPal AI Labs (\href{https://paypal.com}{paypal.com})}

\contribution{Correspondence: \href{mailto:rkaranjai@paypal.com}{\texttt{rkaranjai@paypal.com}}}
\date{November 2026}

\abstract{
Chain-of-thought prompting significantly improves the reasoning capabilities of large language models (LLMs) but incurs high inference costs due to lengthy autoregressive decoding. While previous efficiency approaches reduce costs by discarding tokens or utilizing continuous latent representations, these lossy methods fundamentally sacrifice interpretability and auditability. In this work, we introduce a novel compression framework for intermediate reasoning. We apply Byte-Pair Encoding (BPE) to reasoning corpora to learn a compressed vocabulary of Private Use Area (PUA) tokens, each representing frequent semantic spans. After expanding the LLM's vocabulary, we fine-tune the model via self-distillation to reason natively within this compressed space. Because our encoding is strictly 1:1, our method is lossless. We evaluate our approach on Llama-3.1-8B, Qwen2.5-14B, and Qwen2.5-32B across four diverse benchmarks. Our method reduces total generated tokens by 14.7\% on average with reasoning trace lengths dropping by up to 27.1\%, while incurring a negligible average accuracy cost of 1.6 percentage points. Crucially, this approach is orthogonal to existing lossy methods, allowing practitioners to stack lossless and irreversible compression techniques. We demonstrate that a compressed intermediate language provides a highly efficient computational resource that accelerates inference while fully preserving the transparency of LLM reasoning.
}

\begin{document}
\maketitle

\section{Introduction}

Large Language Models (LLMs) consistently achieve state-of-the-art performance on complex reasoning tasks through the use of Chain-of-Thought (CoT) prompting \cite{weiChainThoughtPrompting2022}. By explicitly generating intermediate steps before arriving at a final answer, models can break down intricate problems in mathematics, logic, and coding into manageable pieces. This explicit reasoning process substantially improves the reliability and accuracy of modern architectures, including the Llama-3.1 and Qwen2.5 series. 
Despite these performance gains, generating long reasoning traces introduces a severe computational bottleneck. Because LLMs decode text autoregressively, every intermediate token produced adds direct latency and increases inference costs. As models are deployed at scale for increasingly complex queries, the sheer length of these intermediate steps makes them highly expensive to serve.

To address this overhead, prior research has investigated various ways to shorten the reasoning process. One prominent direction attempts to internalize the reasoning, forcing the model to compute steps in a continuous latent space rather than outputting text \cite{haoTrainingLargeLanguage2024}. Other approaches focus on dropping tokens directly during generation or pruning the input context \cite{xiaTokenSkipControllableChainofThought2025}. A more intuitive alternative is to simply prompt the model to generate shorter responses, or to force it to reason in a naturally dense language like Chinese, which inherently requires fewer tokens to express complex concepts.

However, these existing strategies introduce critical compromises. Latent reasoning and token-skipping methods fundamentally destroy the readability of the reasoning trace, forcing developers to trade away interpretability and auditability for speed. Furthermore, as our fixed-content \emph{Concept-Normalized Token Rate} (\ctr) analysis reveals, asking a model to use a dense language or to "be concise" does not provide true 1:1 compression. Instead, it unpredictably alters the model's generation policy often causing it to silently skip vital verification steps or omit logic entirely. This highlights a persistent challenge, how can we systematically reduce the number of tokens processed during intermediate reasoning without sacrificing human readability or altering the underlying logical steps?

In this paper, we propose a novel compression method for LLM intermediate reasoning. We apply Byte-Pair Encoding (BPE) to reasoning corpora to identify frequent semantic spans, mapping these redundant sequences to entirely new Private Use Area (PUA) Unicode tokens. 

To validate this approach, we implement a three-stage training pipeline. First, we construct the compressed vocabulary and expand the LLM's embedding matrices, using Fast Vocabulary Transfer to provide an informed initialization. Next, we fine-tune the model via self-distillation, teaching it to naturally output these PUA tokens during reasoning. We conduct systematic empirical evaluations across three models (Llama-3.1-8B, Qwen2.5-14B, and Qwen2.5-32B) on four diverse benchmarks (GSM8K, MATH, StrategyQA, and CommonsenseQA). 

Our findings demonstrate that models can successfully learn to reason in this compressed space. Across our evaluations, this lossless approach reduces total generated tokens by 14.7\% on average, with the reasoning traces themselves dropping by up to 27.1\%, while maintaining accuracy within an average of 1.6 percentage points of the raw baseline. Furthermore, we show that this method is entirely complementary to existing lossy strategies; users can easily apply token-skipping techniques on top of our lossless representations if further compression is required.

Our core contributions are as follows:
\begin{itemize}
  \item We introduce a novel, lossless compression approach for intermediate reasoning that improves LLM generation speed and reduces token cost without sacrificing accuracy or human readability.
  \item We detail a three-stage pipeline to construct a compressed language utilizing Byte-Pair Encoding (BPE) ~\citep{sennrichNeuralMachineTranslation2015} to assign Private Use Area (PUA) tokens, allowing seamless integration via Fast Vocabulary Transfer~\citep{geeFastVocabularyTransfer2024} and self-distillation.
  \item We provide a comprehensive evaluation across three checkpoints (Llama-3.1-8B, Qwen2.5-14B, Qwen2.5-32B) and four datasets (GSM8K, MATH, StrategyQA, CommonsenseQA).
  \item We demonstrate that our compressed reasoning reduces total token counts by an average of 14.7\% (with reasoning-generation lengths dropping by up to 27.1\%), while incurring a negligible average accuracy cost of just 1.6 percentage points.
\end{itemize}

\begin{figure}[!t]
    \centering
    \includegraphics[width=\linewidth]{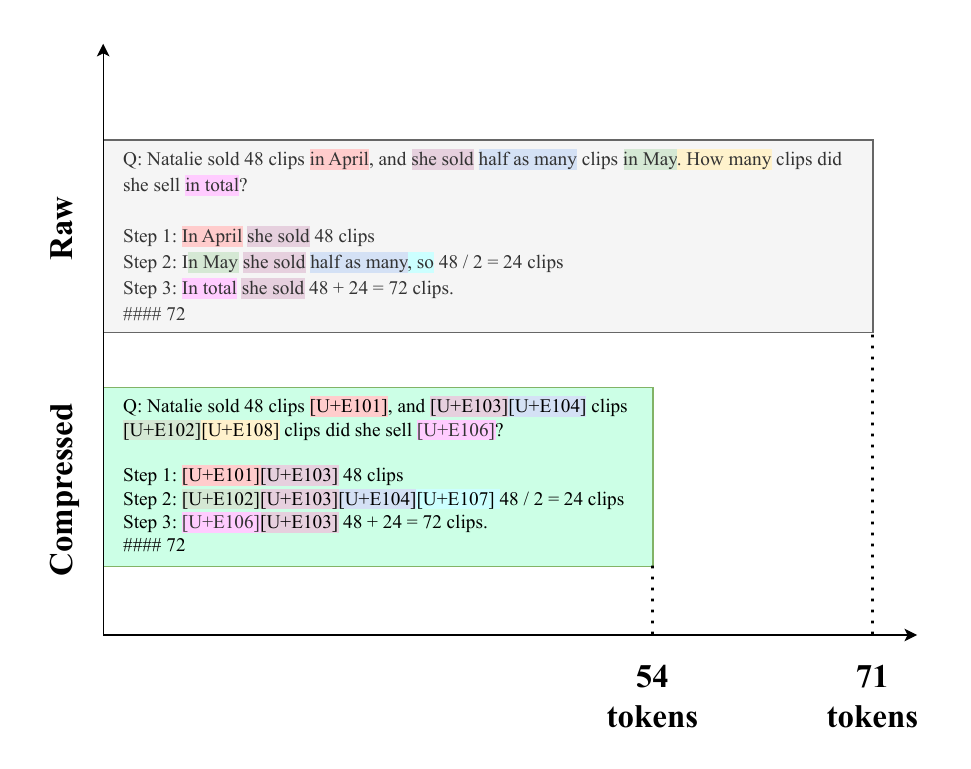}
    \caption{Example of raw vs.\ compressed reasoning on a GSM8K problem. Concept tokens (e.g., \texttt{[U+E103]} $=$ ``she sold'') each replace 2 to 3 words or punctuation marks and can recur across a trace, while the numeric answer is left in plain text so the two versions are graded identically. Here compression reduces the full sequence from 71 to 54 tokens ($-24\%$).}
    \label{fig:1}
\end{figure}

\section{Related Works}\label{sec:related-work}

We review Chain-of-Thought reasoning (Section~\ref{sec:cot}), multilingual and cross-lingual reasoning (Section~\ref{sec:multilingual}), cross-lingual tokenization cost (Section~\ref{sec:tokenization}), and same-language CoT compression (Section~\ref{sec:efficient-cot}), then position our contribution against this landscape (Section~\ref{sec:positioning}).

\subsection{Chain-of-Thought Reasoning}\label{sec:cot}

Chain-of-Thought (CoT) prompting~\citep{weiChainThoughtPrompting2022} elicits a chain of intermediate reasoning steps via few-shot exemplars, substantially improving performance on arithmetic, symbolic, and commonsense reasoning tasks that otherwise show flat scaling curves. \citet{kojimaLargeLanguageModels2022} showed the same effect needs no exemplars at all: appending ``Let's think step by step'' (zero-shot CoT) alone yields large accuracy gains. Reasoning-focused models such as o1 and DeepSeek-R1 extend CoT to thousands of tokens for further gains, but autoregressive decoding makes longer traces roughly linearly more costly~\citep{xiaTokenSkipControllableChainofThought2025}, motivating both the CoT-compression work in Section~\ref{sec:efficient-cot} and the cross-lingual strategy investigated here.

\subsection{Multilingual and Cross-Lingual Reasoning}\label{sec:multilingual}

A separate line of work studies how reasoning quality varies with the reasoning language rather than how to shorten reasoning. \citet{shiLanguageModelsAre2022} introduced MGSM (250 GSM8K problems translated into ten languages) and found CoT ability transfers well to under-represented languages at sufficient model scale. \citet{huangNotAllLanguages2023}'s cross-lingual-thought template (XLT) and \citet{qinCrosslingualPromptingImproving2023}'s cross-lingual (self-consistent) prompting both improve cross-lingual accuracy, by aligning representations across languages or ensembling reasoning paths across them. Across this literature, translating into English, reasoning there, and translating back is a long-standing strong baseline: the same mechanism this project proposes, aimed at English rather than Chinese and motivated by accuracy rather than token cost.

Two findings caution specifically against reasoning in Chinese. \citet{wendlerLlamasWorkEnglish2024} find that LLaMA-2's middle layers pass through an English-centric concept space even for non-English prompts, so changing the surface reasoning language may not change what the model is actually computing internally. \citet{qiWhenModelsReason2025} show, on their XReasoning benchmark, that reasoning models often revert to English when asked to think in another language, and that prompt-based interventions which do force the target language measurably reduce accuracy, a trade-off only partially closed by targeted post-training. Reasoning-language choice therefore carries a real accuracy risk that should not be assumed away.

\subsection{Tokenization Cost and Fairness Across Languages}\label{sec:tokenization}

The premise that Chinese saves tokens rests on an empirical claim about tokenizers. \citet{petrovLanguageModelTokenizers2023} show the same text can require up to 15x more tokens depending on language, a disparity persisting across 17 tokenizers, including multilingual ones. \citet{ahiaAllLanguagesCost2023} show the downstream consequence for commercial APIs: speakers of many languages are simultaneously overcharged and given lower-quality outputs. Together, these results mean token efficiency is a property of the specific tokenizer, not of a language in the abstract, so whether Chinese is cheaper for a given model must be measured directly.

\begin{table}[!ht]
  \centering
  \footnotesize
  \begin{tabular}{@{}l@{\hspace{1ex}}l@{\hspace{1ex}}c@{\hspace{1ex}}c@{\hspace{1ex}}c@{\hspace{1ex}}c@{}}
    Model & Lang & GSM8K & CommonsenseQA & StrategyQA & MATH \\
    & & \shortstack{acc.\\tok.} & \shortstack{acc.\\tok.} & \shortstack{acc.\\tok.} & \shortstack{acc.\\tok.} \\
    \toprule
    \multirow{2}{*}{Qwen-14B}  & EN & \shortstack{91.7\%\\227.5} & \shortstack{82.3\%\\216.3} & \shortstack{76.2\%\\209.4} & \shortstack{69.2\%\\575.0} \\
                                            & ZH & \shortstack{91.0\%\\195.5} & \shortstack{75.8\%\\197.9} & \shortstack{73.8\%\\165.8} & \shortstack{62.8\%\\516.4} \\
    \midrule
    \multirow{2}{*}{Qwen-32B}  & EN & \shortstack{90.8\%\\202.5} & \shortstack{86.5\%\\158.8} & \shortstack{79.4\%\\160.4} & \shortstack{65.4\%\\445.1} \\
                                            & ZH & \shortstack{93.5\%\\198.0} & \shortstack{79.0\%\\222.4} & \shortstack{75.0\%\\190.5} & \shortstack{66.2\%\\454.8} \\
    \midrule
    \multirow{2}{*}{Llama-8B} & EN & \shortstack{82.0\%\\218.5} & \shortstack{77.7\%\\141.0} & \shortstack{68.8\%\\195.3} & \shortstack{39.4\%\\724.1} \\
                                            & ZH & \shortstack{73.3\%\\207.3} & \shortstack{68.6\%\\183.3} & \shortstack{61.9\%\\241.9} & \shortstack{36.2\%\\659.4} \\
  \end{tabular}
  \caption{Baseline accuracy and mean generated tokens on raw English (EN) and Chinese (ZH) prompts.}
  \label{t:baseline-en-zh}
\end{table}

\subsection{Efficient CoT: Compressing Reasoning Within a Language}\label{sec:efficient-cot}

A large body of work instead holds the reasoning language fixed and compresses the trace itself. Some prune or shortcut reasoning steps~\citep{dingBreakChainLarge2024,liuCanLanguageModels2024} or parallelize generation~\citep{ningSkeletonofThoughtLargeLanguage2023}; \citet{marconatoNotAllNeuroSymbolic2023} caution more generally that shortened or re-encoded reasoning can reach high accuracy via unintended shortcuts. Others replace discrete text with continuous computation: pause tokens~\citep{goyalThinkYouSpeak2023}, internalizing steps into hidden states~\citep{dengExplicitCoTImplicit2024}, or reasoning in a continuous latent space~\citep{haoTrainingLargeLanguage2024,chengCompressedChainThought2024}. A third group compresses while remaining in natural language: symbolic notation~\citep{huChainofSymbolPromptingElicits2023}, dynamic token budgets~\citep{hanTokenBudgetAwareLLMReasoning2024}, teacher-distilled short CoTs~\citep{kangC3oTGeneratingShorter2024}, and, closest to our token-level framing, importance-based token pruning~\citep{xiaTokenSkipControllableChainofThought2025}, building on prompt-compression techniques originally developed for inputs rather than reasoning traces~\citep{liCompressingContextEnhance2023,jiangLLMLinguaCompressingPrompts2023,panLLMLingua2DataDistillation2024,wingatePromptCompressionContrastive2022,muLearningCompressPrompts2023,chevalierAdaptingLanguageModels2023,geIncontextAutoencoderContext2024,mohtashamiLandmarkAttentionRandomAccess2023}; closest in spirit to our own natural-language-preserving approach, \citet{chuangLearningCompressPrompt2024} compress prompts into a still human-readable ``Capsule Prompt'' that transfers across black-box LLMs. All of these share one property the present work does not: the reasoning stays in the language it started in, and efficiency comes from re-encoding content rather than changing its language.

\subsection{Positioning}\label{sec:positioning}

To our knowledge, this work is the first to treat reasoning-language choice itself as a lever for token efficiency: translating into Chinese, reasoning there, and translating back. This sits in a genuine gap: the CoT-compression work in Section~\ref{sec:efficient-cot} has not considered cross-lingual recoding; the multilingual-reasoning work in Section~\ref{sec:multilingual} studies cross-lingual prompting almost exclusively toward English and for accuracy rather than token cost, and its accuracy-side evidence is a caution rather than a validation for our premise; and the tokenizer literature in Section~\ref{sec:tokenization} confirms real cross-lingual token disparities without establishing that they favor Chinese for any given model. We therefore treat whether Chinese reasoning saves tokens without an accuracy cost as an open empirical question to be measured directly, not assumed.

\section{Compressed-Language Hypothesis}
Let $x$ be a problem, $m$ a checkpoint, and $s\in\mathcal{S}$ an intermediate language for reasoning. In our paradigm, the prompt and final answer remain in the user's native language; only the intermediate language changes. The model generates a reasoning trace $h$ and a final answer $y$, formulated as:
\[
(h,y)\sim p_m(\cdot\mid x,s).
\]
We record the answer accuracy as $A(m,s)$ and the number of generated completion tokens as $C(m,s)$. Let $C_s=C(m,s)$ and $A_s=A(m,s)$, with ``en'' denoting the standard English reference. A useful compressed intermediate language must satisfy:
\[
\begin{aligned}
\Delta C &= \mathbb{E}[C_s-C_{\mathrm{en}}] < 0,\\
\Delta A &= A_s-A_{\mathrm{en}} \ge -\epsilon,
\end{aligned}
\]
where $\epsilon$ is a prespecified, acceptable accuracy margin. Note that completion length serves as a proxy for decoding cost, not a comprehensive hardware-level metric; it inherently excludes factors such as repeated prefills, latency, energy consumption, and memory overhead.

\subsection{Concept-Normalized Token Rate}
A shorter, freely generated trace may simply omit reasoning steps; therefore, completion length alone cannot prove that a given compressed language represents the same reasoning more efficiently. To isolate representation cost, we define a canonical trace $z=(a_1,\ldots,a_K)$ comprising $K$ typed semantic parts, which include declarations, operations, relations, and answer assertions. A deterministic renderer $r_\ell(z)$ translates and expresses each atom exactly once in language $\ell$. For a given tokenizer $\tau_m$, the \emph{Concept-Normalized Token Rate} (\ctr) is defined as:
\[
\ctr(m,\ell,z)=\frac{|\tau_m(r_\ell(z))|}{K}.
\]
Because the trace $z$ is strictly fixed, a lower \ctr{} cannot result from dropping reasoning content. Crucially, this quantity is jointly indexed by both language and tokenizer; it is not an intrinsic, universal scalar property of any language.

The compact-language hypothesis operates on two distinct levels. The \emph{representation hypothesis} posits that a candidate language will exhibit a lower \ctr{}. Meanwhile, the \emph{realized scratchpad hypothesis} predicts a lower overall generated cost while maintaining acceptable accuracy. While the former serves as a plausible mechanism for the latter, it is not sufficient on its own: a checkpoint's linguistic proficiency and generation policy can amplify, negate, or entirely reverse the underlying representation effect.

\section{Methodology}
Let $\mathcal{M}_\theta$ denote the pretrained LLM with parameters $\theta$, and let $V_o$ be its original (raw) vocabulary, so a raw-language text is a token sequence $x = (x_1, \ldots, x_k) \in V_o^{*}$.

We frame compressed reasoning as a three-part problem:
\begin{enumerate}
	\item First, create a compressed language representation by introducing new tokens.
	\item Next, introduce these new tokens into the model by adding new embeddings to the LLM.
	\item Lastly, fine-tune the model via self-distillation on reasoning data to reason in the compressed space.
\end{enumerate}
In Figure~\ref{fig:1}, we illustrate this with a concrete example: a raw and compressed reasoning trace on the same GSM8K problem, where concept tokens stand in for short recurring spans of text while the final answer stays in plain text for grading.

\subsection{Constructing the Compressed Language}
\label{sec:compressed-language}
 
This is the most critical step. We want to create a new representation that can represent a given corpus $\mathcal{D}_o = \{x^{(1)}, \ldots, x^{(M)}\} \subset V_o^{*}$ using far fewer tokens than the raw text does. This is similar to what we have observed with the Chinese language, where a single token can represent multiple tokens from English; however, the difference is that translating into another language loses information, while what we want here is for the compression to be \emph{lossless}. To achieve this we use Byte-Pair Encoding (BPE)~\citep{sennrichNeuralMachineTranslation2015}, where we repeatedly merge the single most frequent adjacent pair into a new symbol, everywhere it appears, until a pre-defined stopping criterion is met.
 
In practice, we tokenize the corpus $\mathcal{D}_o$ with the base tokenizer, then at each step $t = 1, \ldots, K$ count how often every adjacent pair $(a,b)$ occurs across the corpus (after the first $t-1$ merges have already been applied), and merge whichever pair is most frequent into a new symbol $\tilde{v}_t = \text{merge}(a_t, b_t)$ everywhere it occurs. Note that $a_t$ and $b_t$ can themselves be symbols from an earlier merge step. We keep merging until the top pair frequency drops below a threshold $\tau$, or equivalently once we hit a fixed budget of $K$ merges, which produces the \emph{compressed vocabulary} $V_c = \{\tilde{v}_1, \ldots, \tilde{v}_K\}$.
 
Each new symbol $\tilde{v}_t$ gets its own private-use-area (PUA) Unicode codepoint (\texttt{U+E000}\,--\,\texttt{U+F8FF}, spilling into supplementary PUA planes if $K$ is larger than this range), so $V_c \subset \{\text{U+E000}, \ldots, \text{U+F8FF}\}$. PUA codepoints never collide with ordinary text, so once $\tilde{v}_t$ is registered with the tokenizer it always renders, and is always recovered, as a single token. Putting the raw and compressed vocabularies together gives the extended vocabulary $V = V_o \cup V_c$. The merging process itself defines an encoding map $\mathrm{Enc} : V_o^{*} \to V^{*}$ that swaps in PUA symbols wherever a merged pair occurs, along with a decoding map $\mathrm{Dec} : V^{*} \to V_o^{*}$ that decodes $\tilde{v}_t$ to its raw tokens $\{v_i, v_j, \ldots\} \in V_o$. Write $x^{c} = \mathrm{Enc}(x) \in V^{*}$ for the compressed form of a raw sequence $x$. Applying $\mathrm{Enc}$ to the whole corpus gives the compressed corpus $\mathcal{D}_c = \mathrm{Enc}(\mathcal{D}_o) = \{x^{c,(i)}\}_{i=1}^{M}$, a mix of PUA symbols and whatever original tokens never got merged.
 
\subsection{Adding New Tokens into the Model}
\label{sec:injection}
 
Given $V_c$, we extend the model's vocabulary to $V = V_o \cup V_c$. Let $E_o, W_o \in \mathbb{R}^{|V_o| \times d}$ denote the model's input and output embedding matrices restricted to $V_o$, and let $E_c, W_c \in \mathbb{R}^{|V_c| \times d}$ denote the new rows we need to add for $V_c$. To initialize these new embeddings, we use Fast Vocabulary Transfer (FVT)~\citep{geeFastVocabularyTransfer2024}, where both the input and output embeddings of each new token $\tilde{v}_t$ are initialized as the average of the embeddings of the raw tokens $\{v_i, v_j, \ldots\} \in V_o$ from which it was created. This provides a warm start for the $\tilde{v}_t$ embedding instead of initializing with random values. Next, to learn the new embeddings for these new tokens, we do lightweight fine-tuning, where we freeze all other parameters except $E_c$ and $W_c$ and fine-tune for 1--2 epochs on the compressed corpus $\mathcal{D}_c$, minimizing the standard causal language-modeling loss over this corpus.

\subsection{Compressed-Space Fine-Tuning}
\label{sec:compressed-finetuning}
 
The objective of this stage is not to teach the model to reason, as we assume $\mathcal{M}_\theta$ already reasons competently, but rather to teach it to reason within whichever representation its input is provided in: raw language for a raw-language problem, and the compressed representation for a compressed one. Since we are not adding a new reasoning technique but instead performing reasoning in a new representation space, we opt for self-distillation.
 
For a given reasoning problem (query) $q \in V_o^{*}$ with answer $y$, the frozen model first generates a raw-language chain-of-thought (CoT), $c \sim \mathcal{M}_\theta(\cdot \mid q)$, consisting of reasoning steps $c = (c_1, \ldots, c_T)$ with each $c_t \in V_o^{*}$. Both the query and each reasoning step are then passed through the encoder, giving a compressed query $q^{c} = \mathrm{Enc}(q)$ and a compressed CoT $c^{c} = (c_1^{c}, \ldots, c_T^{c})$, where each $c_t^{c} = \mathrm{Enc}(c_t)$. Collecting these pairs across the full set of problems gives us the self-distillation set $\mathcal{S}^{c}$, where every compressed query is paired with its corresponding compressed CoT. To preserve the model's original reasoning capacity, we fine-tune only a set of LoRA adapters $\phi$ on top of the frozen $\theta$, producing the final fine-tuned model $\mathcal{M}_{\theta+\phi}$.

\begin{table}[ht]
  \centering
  \footnotesize
  \setlength{\tabcolsep}{2pt}
  \begin{tabularx}{\columnwidth}{@{}lrrcX@{}}
    \toprule
    Dataset & Train & Eval & Answer & Question Type \\
    \midrule
    GSM8K                   & 7{,}473  & 1{,}319 & \texttt{\#\textbackslash N} & grade-school arithmetic word problems (multi-step numeric) \\
    MATH                    & 12{,}000 & 500     & \texttt{\textbackslash boxed\{\}}       & competition math (algebra, geometry, number theory) \\
    StrategyQA              & 1{,}443  & 160     & yes/no                                  & multi-hop yes/no needing implicit strategy \\
    CommonsenseQA$^\dagger$ & 9{,}741  & 1{,}221 & 5-way choice                            & multiple-choice commonsense (answer-only, no CoT) \\
    \bottomrule
  \end{tabularx}
  \caption{Reasoning datasets used. GSM8K, MATH and StrategyQA carry genuine chain-of-thought and are used for the self-distillation fine-tune; each also has a parallel Chinese (ZH) eval set of identical size. $^\dagger$CommonsenseQA is answer-only, so it is dropped from the Stage-3 self-distill build and kept for raw-arm evaluation only.}
  \label{t:datasets}
  \vspace{-2ex}
\end{table}
 
\begin{table}[ht]
  \centering
  \footnotesize
  \setlength{\tabcolsep}{2pt}
  \begin{tabular}{@{}llrrr@{}}
    \toprule
    Dataset & Model & Base & Comp. & Red.\% \\
    \midrule
    \multirow{2}{*}{GSM8K}
      & LLaMA-3.1-8B & 1{,}160{,}879 & 943{,}294     & 18.74 \\
      & Qwen2.5      & 1{,}338{,}791 & 1{,}121{,}890 & 16.20 \\
    \midrule
    \multirow{2}{*}{CommonsenseQA}
      & LLaMA-3.1-8B & 230{,}115     & 197{,}000     & 14.39 \\
      & Qwen2.5      & 230{,}229     & 197{,}389     & 14.26 \\
    \midrule
    \multirow{2}{*}{StrategyQA}
      & LLaMA-3.1-8B & 73{,}941      & 66{,}042      & 10.68 \\
      & Qwen2.5      & 76{,}515      & 68{,}637      & 10.30 \\
    \midrule
    \multirow{2}{*}{MATH}
      & LLaMA-3.1-8B & 3{,}371{,}036 & 2{,}844{,}944 & 15.61 \\
      & Qwen2.5      & 3{,}519{,}358 & 2{,}993{,}423 & 14.94 \\
    \midrule
    \multirow{2}{*}{\textbf{Overall}}
      & LLaMA-3.1-8B & 4{,}835{,}971 & 4{,}051{,}280 & \textbf{16.23} \\
      & Qwen2.5      & 5{,}164{,}893 & 4{,}381{,}339 & \textbf{15.17} \\
    \bottomrule
  \end{tabular}
  \caption{Per-dataset compression of the English training corpus under each model's concept codec. \emph{base}: base-tokenizer tokens; \emph{comp}: compressed units; \emph{red.\%} $=100(1-\text{comp}/\text{base})$. GSM8K compresses most; StrategyQA least.}
  \label{t:compression}
  \vspace{-2ex}
\end{table}
 
\begin{table}[ht]
  \centering
  \footnotesize
  \setlength{\tabcolsep}{2pt}
  \begin{tabular}{@{}lrr@{}}
    \toprule
    Concept Tokens & LLaMA-3.1-8B & Qwen2.5 \\
    \midrule
    GSM8K          & 1{,}616   & 1{,}593 \\
    CommonsenseQA  & 857       & 841 \\
    StrategyQA     & 717       & 707 \\
    MATH           & 2{,}794   & 2{,}773 \\
    \midrule
    distinct EN    & \textbf{2{,}944} & \textbf{2{,}919} \\
    base vocab.    & 128{,}256 & 151{,}665 \\
    ext. vocab. (EN) & 131{,}200 & 154{,}584 \\
    \bottomrule
  \end{tabular}
  \caption{English concept (PUA) tokens added to the vocabulary. The first four rows are the number of distinct concept tokens occurring in each dataset's English corpus; these overlap across datasets, so they exceed the distinct EN total. MATH alone exercises $\sim$95\% of the learned codec. ``ext. vocab. (EN)'' $=$ base vocab. $+$ distinct EN concepts.}
  \label{t:vocab}
  \vspace{-2ex}
\end{table}

\section{Evaluation and Results}

Before evaluating mixed-language intermediate reasoning, we first establish the baseline monolingual competence of our selected checkpoints. Table~\ref{t:baseline-en-zh} reports the accuracy and mean token counts when both the prompt and the expected reasoning are presented entirely in either English (EN) or Chinese (ZH). While the Qwen checkpoints maintain competitive accuracy across both languages, confirming their viability for our compressed-language intervention, Llama-3.1-8B exhibits a clear baseline degradation in Chinese (e.g., dropping from 82.0\% to 73.3\% on GSM8K). This baseline capability gap neatly anticipates our later findings: a compressed intermediate language can only yield efficient reasoning if the underlying checkpoint possesses robust native competence in that language.

We evaluate our approach across multiple reasoning datasets and multiple models: Llama-3.1-8B-Instruct, Qwen2.5-14B-Instruct, and Qwen2.5-32B-Instruct. The datasets we use are described in Table~\ref{t:datasets}.
 
For each dataset, we take its training split as the raw corpus $\mathcal{D}_o$ and learn the compressed vocabulary $V_c$ from it as described in Section~\ref{sec:compressed-language}. We restrict these experiments to English text, using the Qwen tokenizer as the base tokenizer for the Qwen models and the LLaMA tokenizer for the LLaMA models, and we stop merging once the highest pair frequency falls below the threshold $\tau = 5$ after $K$ merge steps. The resulting compressed-language statistics are given in Table~\ref{t:compression} and Table~\ref{t:vocab}. Because we serve every model as a batch in vLLM, wall-clock latency is not a reliable point of comparison across raw and compressed configurations; we instead measure efficiency directly in tokens, i.e., the number of tokens saved, or equivalently the number of fewer tokens the model must process as a prompt or produce as a generation. This is the efficiency metric reported throughout in  Table~\ref{t:en-combined-both-models}.
 
When constructing $V_c$, we exclude mathematical symbols, \LaTeX{} symbols, and numbers from merging, leaving them as ordinary tokens from $V_o$. Including them would introduce new PUA tokens for mathematical notation, which would require substantially more training for the model to learn to reason correctly with them.
 
To build the reasoning fine-tuning data (Section~\ref{sec:compressed-finetuning}), we use GSM8K, MATH, and StrategyQA; we do not use CommonsenseQA, since it has no chain-of-thought to distill. For the embedding warm-start stage (Section~\ref{sec:injection}), we train on the compressed corpus $\mathcal{D}_c$ for 2 epochs on 3 H100 GPUs, with every parameter frozen except the new embedding rows $E_c$ and $W_c$.
 
For the self-distillation stage (Section~\ref{sec:compressed-finetuning}), we generate raw-language reasoning steps $c$ for each query $q$ in the training split of GSM8K, MATH, and StrategyQA, then apply our lossless compression to obtain the compressed pairs $(q^c, c^c)$ that make up $\mathcal{S}^c$. We fine-tune the LoRA adapters $\phi$ on $\mathcal{S}^c$ for 3--4 epochs on 3 H100 GPUs.

\begin{table*}[!ht]
  \centering
  \footnotesize
  \begin{tabular}{@{}l@{\hspace{1ex}}l|c|c@{\hspace{1ex}}c@{\hspace{1ex}}c@{\hspace{2ex}}|r@{\hspace{1ex}}r@{\hspace{1ex}}r@{\hspace{1ex}}r@{}}
    Dataset & Key & Acc. (\%) & \multicolumn{3}{c|}{Both Correct} & \multicolumn{4}{c}{Total (All Problems)} \\
    & & & $n$ & Gen & Total & Prompt & Gen. & Total & \shortstack{Gen.\\PUA\%} \\
    \toprule
    \multicolumn{10}{c}{\textbf{Qwen2.5-14B-Instruct}} \\
    \midrule
    GSM8K         & \shortstack{Raw\\Compressed\\\% token saved} & \shortstack{\textbf{90.2\%}\\89.4\%} & 1128 & \shortstack{213.2\\\textbf{164.9}\\22.7\%} & \shortstack{374.9\\\textbf{306.1}\\18.3\%} & \shortstack{163.3\\\textbf{142.7}\\12.6\%} & \shortstack{221.4\\\textbf{172.1}\\22.3\%} & \shortstack{384.7\\\textbf{314.7}\\18.2\%} & \shortstack{0.0\%\\22.9\%} \\
    \midrule
    MATH          & \shortstack{Raw\\Compressed\\\% token saved} & \shortstack{\textbf{67.8\%}\\67.4\%} & 311  & \shortstack{444.4\\\textbf{373.5}\\16.0\%} & \shortstack{593.6\\\textbf{507.4}\\14.5\%} & \shortstack{166.2\\\textbf{149.3}\\10.2\%} & \shortstack{548.5\\\textbf{457.5}\\16.6\%} & \shortstack{714.6\\\textbf{606.8}\\15.1\%} & \shortstack{0.0\%\\13.1\%} \\
    \midrule
    StrategyQA    & \shortstack{Raw\\Compressed\\\% token saved} & \shortstack{78.7\%\\78.7\%}          & 120  & \shortstack{187.1\\\textbf{158.9}\\15.1\%} & \shortstack{301.2\\\textbf{265.3}\\11.9\%} & \shortstack{114.4\\\textbf{106.7}\\6.7\%} & \shortstack{199.4\\\textbf{167.0}\\16.3\%} & \shortstack{313.8\\\textbf{273.7}\\12.8\%} & \shortstack{0.0\%\\13.6\%} \\
    \midrule
    CommonsenseQA & \shortstack{Raw\\Compressed\\\% token saved} & \shortstack{\textbf{82.5\%}\\81.6\%} & 937  & \shortstack{228.2\\\textbf{176.9}\\22.4\%} & \shortstack{367.5\\\textbf{303.4}\\17.5\%} & \shortstack{139.5\\\textbf{126.5}\\9.3\%} & \shortstack{233.6\\\textbf{183.9}\\21.3\%} & \shortstack{373.1\\\textbf{310.4}\\16.8\%} & \shortstack{0.0\%\\12.7\%} \\
    \toprule
    \multicolumn{10}{c}{\textbf{Meta-Llama-3.1-8B-Instruct}} \\
    \midrule
    GSM8K         & \shortstack{Raw\\Compressed\\\% token saved} & \shortstack{\textbf{82.3\%}\\78.6\%} & 949  & \shortstack{192.1\\\textbf{140.1}\\27.1\%} & \shortstack{322.2\\\textbf{260.1}\\19.3\%} & \shortstack{133.0\\\textbf{122.3}\\8.1\%} & \shortstack{214.5\\\textbf{164.0}\\23.5\%} & \shortstack{347.4\\\textbf{286.3}\\17.6\%} & \shortstack{0.0\%\\25.4\%} \\
    \midrule
    MATH          & \shortstack{Raw\\Compressed\\\% token saved} & \shortstack{38.0\%\\\textbf{39.4\%}} & 144  & \shortstack{330.5\\\textbf{260.3}\\21.2\%} & \shortstack{440.6\\\textbf{366.6}\\16.8\%} & \shortstack{135.8\\\textbf{129.0}\\5.0\%} & \shortstack{698.3\\\textbf{593.8}\\15.0\%} & \shortstack{834.1\\\textbf{722.7}\\13.4\%} & \shortstack{0.0\%\\16.0\%} \\
    \midrule
    StrategyQA    & \shortstack{Raw\\Compressed\\\% token saved} & \shortstack{\textbf{75.0\%}\\72.5\%} & 106  & \shortstack{130.7\\\textbf{122.8}\\6.0\%} & \shortstack{274.6\\\textbf{256.9}\\6.4\%} & \shortstack{144.2\\\textbf{134.5}\\6.7\%} & \shortstack{153.4\\\textbf{134.3}\\12.5\%} & \shortstack{297.6\\\textbf{268.7}\\9.7\%} & \shortstack{0.0\%\\11.2\%} \\
    \midrule
    CommonsenseQA & \shortstack{Raw\\Compressed\\\% token saved} & \shortstack{\textbf{76.7\%}\\73.1\%} & 818  & \shortstack{145.8\\\textbf{138.6}\\4.9\%} & \shortstack{304.5\\\textbf{283.3}\\7.0\%} & \shortstack{158.5\\\textbf{144.4}\\8.9\%} & \shortstack{155.0\\\textbf{151.5}\\2.3\%} & \shortstack{313.4\\\textbf{295.9}\\5.6\%} & \shortstack{0.0\%\\12.0\%} \\
    \toprule
    \multicolumn{10}{c}{\textbf{Qwen2.5-32B-Instruct}} \\
    \midrule
    GSM8K         & \shortstack{Raw\\Compressed\\\% token saved} & \shortstack{\textbf{91.7\%}\\90.3\%} & 1151 & \shortstack{186.3\\\textbf{145.7}\\21.8\%} & \shortstack{348.0\\\textbf{287.0}\\17.5\%} & \shortstack{163.3\\\textbf{142.7}\\12.6\%} & \shortstack{193.6\\\textbf{150.4}\\22.3\%} & \shortstack{356.9\\\textbf{293.1}\\17.9\%} & \shortstack{0.0\%\\21.8\%} \\
    \midrule
    MATH          & \shortstack{Raw\\Compressed\\\% token saved} & \shortstack{\textbf{63.0\%}\\60.4\%} & 270  & \shortstack{335.8\\\textbf{261.2}\\22.2\%} & \shortstack{483.2\\\textbf{393.5}\\18.6\%} & \shortstack{166.2\\\textbf{149.3}\\10.2\%} & \shortstack{424.4\\\textbf{320.9}\\24.4\%} & \shortstack{590.6\\\textbf{470.2}\\20.4\%} & \shortstack{0.0\%\\21.6\%} \\
    \midrule
    StrategyQA    & \shortstack{Raw\\Compressed\\\% token saved} & \shortstack{81.2\%\\\textbf{81.9\%}} & 123  & \shortstack{156.8\\\textbf{134.6}\\14.2\%} & \shortstack{271.0\\\textbf{241.0}\\11.1\%} & \shortstack{114.4\\\textbf{106.7}\\6.7\%} & \shortstack{165.1\\\textbf{142.0}\\14.0\%} & \shortstack{279.6\\\textbf{248.7}\\11.1\%} & \shortstack{0.0\%\\13.8\%} \\
    \midrule
    CommonsenseQA & \shortstack{Raw\\Compressed\\\% token saved} & \shortstack{\textbf{86.2\%}\\84.7\%} & 1001 & \shortstack{165.9\\\textbf{122.6}\\26.1\%} & \shortstack{305.5\\\textbf{249.2}\\18.4\%} & \shortstack{139.5\\\textbf{126.5}\\9.3\%} & \shortstack{170.7\\\textbf{127.2}\\25.5\%} & \shortstack{310.2\\\textbf{253.6}\\18.3\%} & \shortstack{0.0\%\\13.6\%} \\
  \end{tabular}
  \caption{Comparison across models: raw vs.\ compressed accuracy, the matched both-correct subset, and full totals across all problems. Each dataset is a single row; the \emph{key} column repeats the stacking order used by every token-count cell (\emph{raw} on top, \textbf{compressed} in bold below, \% tokens saved on the bottom line). Accuracy and genPUA\% cells stack only raw/compressed, with no \% line. Bold marks the better value in each pair: for accuracy, the higher percentage wins; for token counts, the lower count wins.}
  \label{t:en-combined-both-models}
  \vspace{-3.5ex}
\end{table*}

Table~\ref{t:en-combined-both-models} demonstrates the validity of our hypothesis that compressing the input losslessly, and teaching the model to reason directly in that compressed space, is easy to do and substantially reduces the number of tokens the model has to process, at a very small cost in accuracy.
 
Averaged across all twelve model/dataset rows in Table~\ref{t:en-combined-both-models} (three models, four datasets each), compressed reasoning cuts total token counts by \textbf{14.7\% on average} (5.6--20.4\% depending on model and dataset), while accuracy moves by only \textbf{1.6 percentage points on average}. It stays under one point in five of the twelve rows, and never falls by more than 3.7 points even in the worst case (Llama-8B on GSM8K, $82.3\% \to 78.6\%$). Compressed reasoning matches raw exactly on one row (Qwen-14B on StrategyQA, $78.7\% \to 78.7\%$) and slightly beats it on two others (Llama-8B on MATH, $38.0\% \to 39.4\%$; Qwen-32B on StrategyQA, $81.2\% \to 81.9\%$). So the accuracy cost is not just small on average, but for some dataset/model pairs it is effectively absent.
 
The gap is even larger once we isolate the effect on reasoning length itself. On the matched both-correct subset, where raw and compressed answer the same problems correctly, generation length alone drops by 4.9--27.1\% across the twelve rows (mean $\approx 18\%$), confirming that most of the savings in Table~\ref{t:en-combined-both-models} come from shorter reasoning traces rather than shorter prompts. The genPUA\% column shows the model is genuinely reasoning in the compressed vocabulary rather than merely compressing its input: 11.2--25.4\% of generated tokens are concept (PUA) tokens in the compressed arm, compared with $0.0\%$ by construction in the raw arm.
 
We achieve this even with a small codec: only 2{,}944 (LLaMA) and 2{,}919 (Qwen) new concept tokens on top of base vocabularies of well over 100{,}000 tokens. Learning this codec from a larger corpus, with a larger merge budget $K$, should recover proportionally more redundancy, translating directly into fewer tokens to process at every subsequent reasoning step.
 
Because $\mathrm{Enc}$ and $\mathrm{Dec}$ are exact inverses (Section~\ref{sec:compressed-language}), every compressed reasoning trace we generate can always be losslessly decoded back into ordinary, human-readable text. This is a meaningful advantage over efficiency methods that shorten reasoning by discarding it outright. Coconut~\citep{haoTrainingLargeLanguage2024}, for example, replaces chain-of-thought tokens with continuous latent vectors that are never decoded back into language at all, permanently trading away the ability to read, audit, or debug the model's reasoning. Our compressed vocabulary $V_c$ never destroys information; it only gives the model a shorter, learned name for spans of text it already produces. Nothing about the underlying reasoning becomes unrecoverable or opaque, unlike methods that permanently remove text from the chain of thought.
 
CommonsenseQA is excluded from the self-distillation fine-tune (Table~\ref{t:datasets}, $^\dagger$), but it still shows meaningful compression and PUA usage at evaluation time (12.7\% and 13.6\% genPUA\% for Qwen-14B and Qwen-32B, 12.0\% for Llama-8B). This is evidence that the model learned a general \emph{reason in whichever space the input is given} behavior rather than memorizing per-dataset shortcuts. The trade-off is not uniform, though. Llama-8B pays the largest accuracy cost on GSM8K and CommonsenseQA ($3.7$ and $3.6$ percentage points) while banking the smallest token savings in the whole table on CommonsenseQA specifically ($5.6\%$), suggesting Llama-8B generalizes to compressed-language reasoning less cleanly than either Qwen model in this setup, which could be due to the bigger size of the Qwen models, as larger models can adopt a new compressed language more effectively.

\section{Discussion}

Our experiments support the core premise of the compressed-language hypothesis: intermediate reasoning can be significantly shortened without sacrificing accuracy, provided the representation is formally constructed rather than borrowed from natural language. As our fixed-content \ctr\ analysis established, attempting to use Chinese or concise-English as an intermediate language achieves token savings primarily by unpredictably altering the model's reasoning policy (omitting steps), rather than through true 1:1 representation efficiency. By contrast, our lossless BPE-PUA compression method reliably reduces total generated tokens by an average of 14.7\%, and reasoning-trace length by up to 27.1\%, by genuinely increasing the information density of the tokens themselves. 

Crucially, our results demonstrate that successful adoption of this compressed representation depends on both the architecture's capacity and its baseline competence. For example, Qwen2.5-14B and Qwen2.5-32B seamlessly adapted to the PUA vocabulary across all tasks, generating 11.2\% to 26.1\% of their trace using the new concept tokens while incurring near-zero accuracy penalties. Qwen2.5-32B even saw a slight accuracy improvement on StrategyQA ($81.2\% \to 81.9\%$). Conversely, Llama-3.1-8B struggled to generalize as cleanly; it suffered the largest accuracy drops (e.g., $82.3\% \to 78.6\%$ on GSM8K) and banked the smallest token savings on CommonsenseQA. This aligns with our baseline monolingual evaluation: models that possess inherently stronger representation capacities (typically larger parameters) can more effectively remap complex reasoning logic into novel, compressed embedding spaces.

Furthermore, the results on CommonsenseQA, which was explicitly excluded from the self-distillation fine-tuning phase, reveal a critical behavior: the models did not simply memorize compressed reasoning shortcuts for specific datasets. Instead, they learned a general heuristic to \emph{reason in whichever representational space the input is provided}. This strong out-of-domain generalization confirms that lossless compression fundamentally alters the model's token-emission efficiency without compromising its underlying logical framework.

Our approach cleanly disentangles two properties that lossy compression methods frequently conflate: \emph{sequence compactness} and \emph{reasoning exactness}. Unlike latent-space internalization or irreversible token-skipping, our compressed vocabulary never destroys information. The reasoning trace remains fully auditable and human-readable. This proves that output compactness need not come at the expense of interpretability; LLMs can successfully learn to reason efficiently within highly compressed, yet perfectly lossless, intermediate substrates.

\section{Limitations and Broader Impact}

Our study isolates the efficiency of intermediate reasoning languages by maintaining prompts and answers in English; evaluating the full multilingual user experience requires independently varying source and output languages. Methodologically, our fixed-content evaluation relies on a single deterministic renderer. Future work must incorporate native-speaker paraphrases and strictly track language compliance to account for code-switching or unsolicited English fallback. Furthermore, completion tokens serve strictly as a decode-length proxy. Establishing definitive hardware-normalized efficiency claims requires comprehensive accounting of active latency, memory footprint, and KV-cache usage. Our exploratory router also relies on a single data split, and Qwen3 was evaluated using deterministic decoding; robust deployment guarantees necessitate validation-locked thresholds and repeated evaluations. To support reproducibility, our anonymous artifact includes complete item-level generations, prompts, tokenizer hashes, and router splits.

\textbf{AI Assistance Disclosure.} AI models were used to assist with writing and debugging code. All AI-assisted outputs were reviewed, verified, and edited by the authors, who take full responsibility for the content, code, and results presented in this paper.

\section{Conclusion}
In this work we present a novel approach to reducing the number of tokens that an LLM has to process during reasoning, thus improving LLM reasoning speed without compromising on the accuracy. We also demonstrate that by using a simple lossless compression, in which every compressed reasoning trace we produce can always be decoded back into ordinary, human-readable text, we can still substantially reduce the reasoning cost. This does not preclude combining our method with orthogonal, lossy techniques, where users who are willing to trade away interpretability for savings beyond what lossless compression alone provides can still stack a method such as TokenSkip, which shortens a chain of thought by irreversibly skipping less important tokens, on top of our approach to reduce cost further.

\bibliographystyle{plainnat}
\bibliography{references}

\end{document}